\documentclass{article}
\usepackage{arxiv}

\usepackage[utf8]{inputenc} % allow utf-8 input
\usepackage[T1]{fontenc}    % use 8-bit T1 fonts
\usepackage{hyperref}       % hyperlinks
\usepackage{url}            % simple URL typesetting
\usepackage{booktabs}       % professional-quality tables
\usepackage{amsfonts}       % blackboard math symbols
\usepackage{nicefrac}       % compact symbols for 1/2, etc.
\usepackage{microtype}      % microtypography
\usepackage{lipsum}
\usepackage{graphicx}
\graphicspath{ {./images/} }
\usepackage{booktabs}  % Add this in your preamble if not already included
\usepackage{array}
\usepackage[round]{natbib} 
\usepackage{xcolor}
\usepackage{caption}
\usepackage{hyperref}
\usepackage{cleveref}
\crefname{figure}{Fig.}{Figs.}
\hypersetup{
    colorlinks=true,
    linkcolor=blue,    % for \ref, \pageref, TOC
    citecolor=blue,    % for \cite
    urlcolor=blue      % for \url
}

\newcommand{\RN}[1]{\uppercase\expandafter{\romannumeral#1}}

\title{A Novel Multi-Agent Architecture to Reduce Hallucinations of Large Language Models in Multi-step Structural Modeling}

\author{
Ziheng Geng$^{1,*}$,
Jiachen Liu$^{2,*}$,
Ran Cao$^{3}$,
Lu Cheng$^{4}$,
Dan M. Frangopol$^{5}$,
Minghui Cheng$^{1,6\dagger}$\\
\\
$^{1}$Department of Civil and Architectural Engineering, University of Miami, Coral Gables, FL 33146, USA\\
$^{2}$HBC Engineering Company, Miami, FL 33178, USA\\
$^{3}$College of Civil Engineering, Hunan University, Changsha, 410082, China\\
$^{4}$Department of Computer Science, University of Illinois Chicago, Chicago, IL 60607, USA\\
$^{5}$Department of Civil and Environmental Engineering, Lehigh University, Bethlehem, PA 18015, USA\\
$^{6}$School of Architecture, University of Miami, Coral Gables, FL 33146, USA\\
\\
$^{*}$Equal contribution.\\
$^{\dagger}$Corresponding author: \texttt{minghui.cheng@miami.edu}
}

\begin{document}
\maketitle
\begin{abstract}
% Recent advances in large language models (LLMs) have empowered multi-agent architectures to address complex engineering challenges through collaborative intelligence. Nevertheless, the coordination, connectivity, and communication among agents within these systems remain largely unexplored, despite its critical role in ensuring accuracy, efficiency, and robustness. This study presents a novel architectural design for LLM-based multi-agent architectures to automate the structural analysis of 2D frames using OpenSeesPy. The workflow begins with problem analysis and construction planning agents that extract key parameters from user descriptions and formulate a stepwise construction plan. Next, node and element agents operate in parallel to assemble the frame geometry, followed by a load assignment agent that applies nodal and elemental loads. The resulting geometric and load data are then converted into executable OpenSeesPy script by two code translation agents. The system employs two backbone LLMs: GPT-OSS 120B for complex reasoning and Llama-3.3 70B Instruct Turbo for information mapping and transformation. System performance is evaluated on a benchmark of 20 frame problems over ten repeated trials, achieving 100\% accuracy in 18 cases and 90\% in the remaining two. Additionally, compared to sequential multi-agent designs, the proposed architecture significantly improves computational efficiency and demonstrates scalability to larger structural systems.

Large language models (LLMs) such as GPT and Gemini have demonstrated remarkable capabilities in contextual understanding and reasoning. The strong performance of LLMs has sparked growing interest in leveraging them to automate tasks traditionally dependent on human expertise. Recently, LLMs have been integrated into intelligent agents capable of operating structural analysis software (e.g., OpenSees) to construct structural models and perform analyses. However, existing LLMs are limited in handling multi-step structural modeling due to frequent hallucinations and error accumulation during long-sequence operations. To this end, this study presents a novel multi-agent architecture to automate the structural modeling and analysis using OpenSeesPy. First, problem analysis and construction planning agents extract key parameters from user descriptions and formulate a stepwise modeling plan. Node and element agents then operate in parallel to assemble the frame geometry, followed by a load assignment agent. The resulting geometric and load information is translated into executable OpenSeesPy scripts by code translation agents. The proposed architecture is evaluated on a benchmark of 20 frame problems over ten repeated trials, achieving 100\% accuracy in 18 cases and 90\% in the remaining two. The architecture also significantly improves computational efficiency and demonstrates scalability to larger structural systems.
\end{abstract}

% keywords can be removed
\begin{quote}
\end{quote}

\section{Introduction}
Structural analysis is a fundamental pillar of civil engineering, underpinning the design and evaluation of buildings and infrastructure to ensure their safety, stability, and serviceability. Over the past several decades, finite element modeling has emerged as the dominant approach for conducting structural analysis due to its accuracy, versatility, and broad applicability. A variety of commercial and open-source software platforms, such as OpenSees \citep{mckenna2011opensees}, ETABS \citep{etabs2023}, SAP2000 \citep{sap2000}, ANSYS \citep{ansys}, and Abaqus \citep{abaqus}, have been widely adopted in both academia and industry. Despite these technical advances, structural modeling using finite element software is multi-step and remains highly manual and labor-intensive. Engineers are required to perform a series of repetitive steps, such as defining nodes and elements, assigning material properties, applying load patterns, and specifying boundary conditions. These tasks are typically executed through graphical user interfaces (GUIs), which depend on click-based operations and demand considerable domain expertise. Such manual workflows hinder modeling efficiency, underscoring the need for automation in structural analysis.

Trained on massive text datasets and consisting of billions of parameters, large language models (LLMs) such as GPT \citep{openai2025gpt52update} and Gemini \citep{google2025gemini3procard} have demonstrated remarkable capabilities in contextual understanding \citep{an2024make,zhu2024can}, logical reasoning \citep{cheng2025empowering, xie2025logic}, and instruction following \citep{zhou2023instruction, zeng2023evaluating}. These breakthroughs have sparked growing interest in leveraging LLMs to automate tasks traditionally dependent on human expertise. Within the structural engineering community, initial efforts have been made to evaluate LLM’s structural engineering ability. \cite{wan2025som} established a dataset of strength of materials problems and found that general-purpose LLMs are not accurate in solving these problems. To overcome these challenges, supervised fine-tuning and retrieval-augmented generation are employed to integrate domain knowledge. Successful applications include textual interpretations of structural damage images and building surface defects \citep{jiang2025large,xutwo}, generation of construction inspection reports \citep{pu2024autorepo,wang2025integrated}, and information query of building codes and standards \citep{joffe2025framework,shi2025fine}. Another major direction lies in the development of agents. Specifically, the LLM serves as the central reasoning engine that plans a sequence of operations and utilizes computational tools to complete complex engineering tasks. In existing studies, agents have been developed to use Revit to generate and review building models \citep{du2024text2bim,deng2025bimgent,dong2025ai} and to use OpenSeesPy to automate the analysis of beams and frame structures \citep{liu2026large,geng2025lightweight,liang2025integrating}.

The aforementioned studies pioneer the application of LLMs in structural engineering and demonstrate their potential to automate and accelerate existing workflows. However, they remain limited in handling multi-step structural modeling tasks. While supervised fine-tuning and retrieval-augmented generation enable LLMs to acquire qualitative domain knowledge, they lack the quantitative reliability required for rigorous structural calculations \citep{liu2026large}. Agents that combine the qualitative reasoning capabilities of LLMs with the numerical precision of established software are therefore more suitable for structural modeling and analysis. Nevertheless, existing agent frameworks are typically restricted to tasks involving only a small number of steps, whereas realistic structural analysis problems often require hundreds to thousands of steps to construct a valid structural model. As the number of steps increases, error accumulation and hallucination become increasingly severe, leading to invalid structural models. This issue is particularly critical in structural engineering, where even minor errors can compromise safety. Therefore, a new agent architecture is needed to scale long-horizon structural modeling while maintaining consistency, numerical accuracy, computational efficiency, and reliability against hallucination.

This paper proposes a novel multi-agent architecture to reliably and efficiently perform multi-step structural modeling. The architecture decomposes the modeling workflow into four coordinated modules: analysis and planning, geometry assembly, load integration, and code translation. Specifically, problem analysis and construction planning agents first extract key parameters from user input and formulate a sequential modeling plan. The frame geometry is then constructed through parallel node and element agents, after which a load assignment agent applies the corresponding nodal and elemental loads. Finally, code translation agents transform the geometric and loading information into executable OpenSeesPy scripts. The results indicate that the proposed architecture consistently achieves accuracy exceeding 90\% across 20 benchmark problems and scale to larger building structures. The remainder of the paper is organized as follows. Section 2 introduces the benchmark dataset comprising 20 representative frame problems and identifies the limitations of the existing architecture. Section 3 presents the proposed multi-agent architecture, while section 4 shows the evaluation results. Finally, Sections 5 and 6 discuss the study’s limitations and provide the concluding remarks.

\section{Benchmark Dataset and Evaluation}
\label{sec:headings}

\subsection{Dataset overview}
\begin{figure*}[htbp]
\centering
% \captionsetup{justification=centering}
\includegraphics[width=0.8\textwidth]{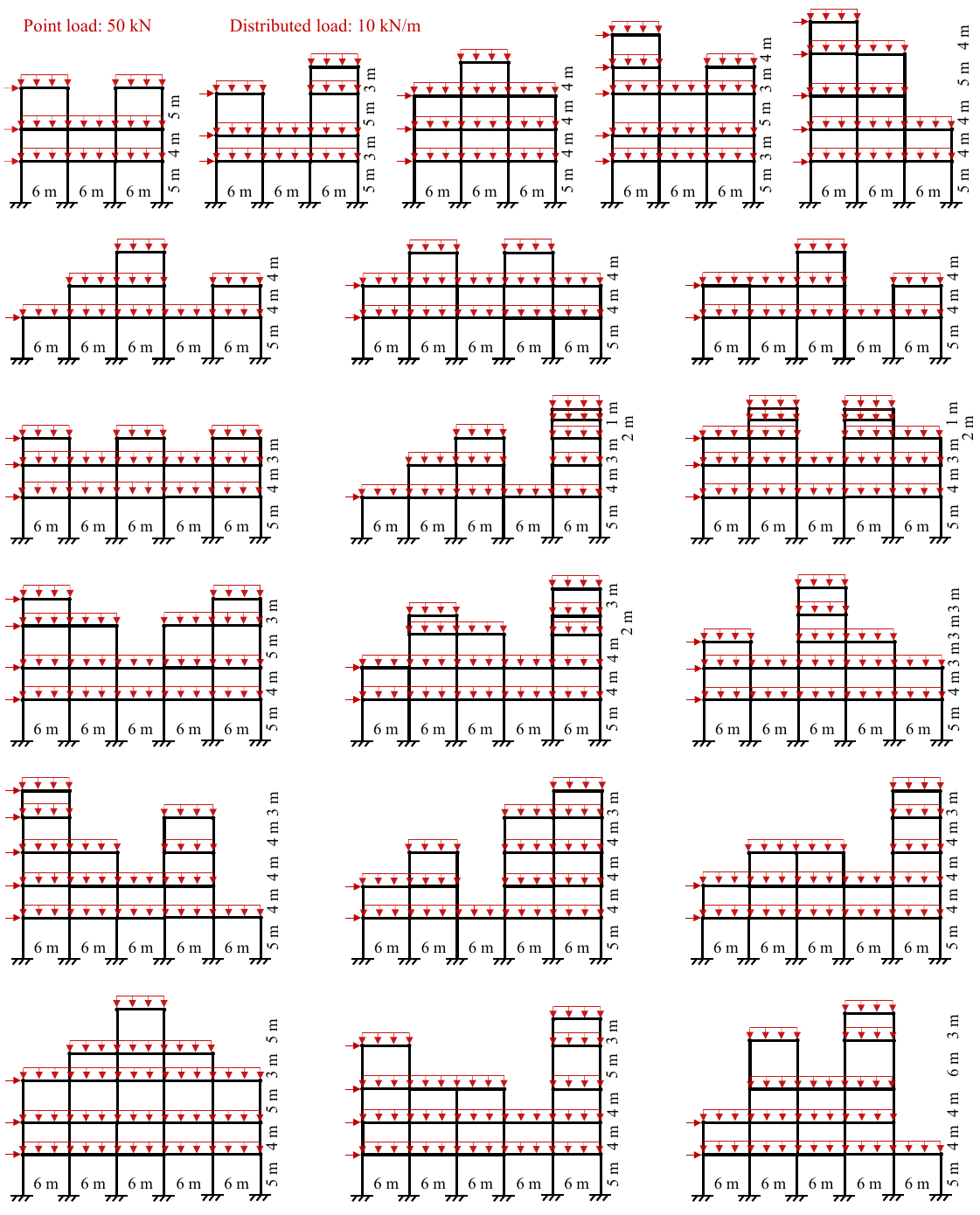}
\caption{Benchmark dataset of twenty frame structural modeling problems (adapted from \citeauthor{geng2025lightweight}, \citeyear{geng2025lightweight}).}
\label{Figure1}
\end{figure*}

A benchmark dataset developed in previous work \citep{geng2025lightweight} is adopted to evaluate the performance of multi-agent LLMs for automated frame structural modeling. The dataset comprises 20 representative frame problems, as illustrated in \cref{Figure1}. Each problem features a unique structural geometry composed of vertical columns and horizontal girders. Among them, five frames have three bays and fifteen contain five bays, where each bay spans six meters. The number of stories in each bay and story heights are randomly sampled between one and five to assess the generalizability. The boundary conditions, load patterns, and material properties are consistent across all problems. Specifically, all supports are fixed at the base. A uniformly distributed load of 10 kN/m is applied downward on each girder, while a point load of 50 kN is applied rightward to the top node of each story at the leftmost bay. The material is specified by three parameters: Young’s modulus, cross-sectional area, and moment of inertia. Distinct cross-sectional properties are assigned to the columns and girders, respectively. These problems provide a systematic testbed for assessing the capability of multi-agent LLMs in solving frame structural modeling. 

\begin{figure*}[htbp]
\centering
% \captionsetup{justification=centering}
\includegraphics[width=0.8\textwidth]{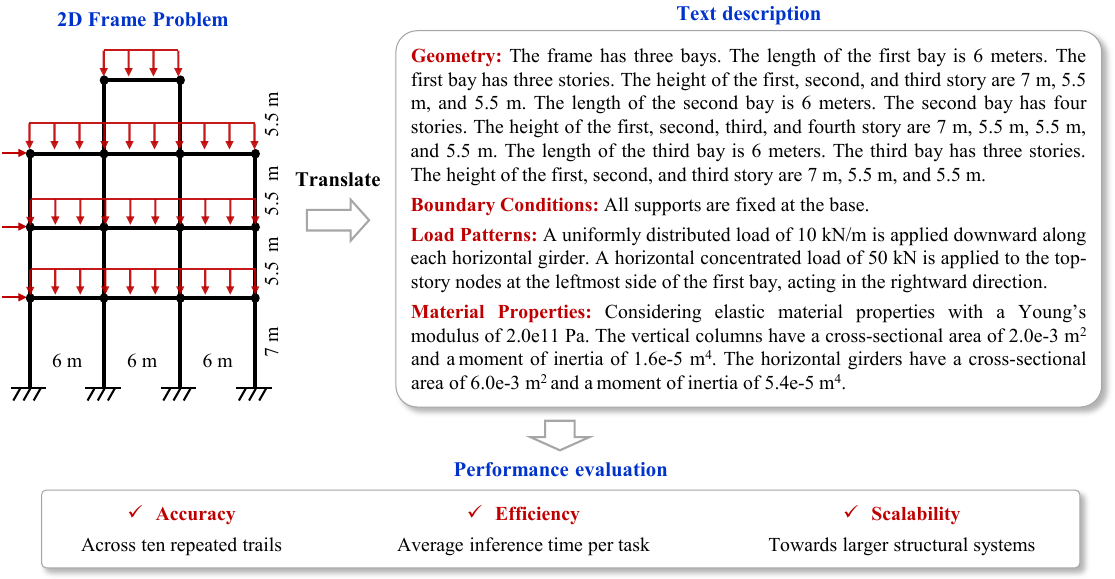}
\caption{Textual description template for specifying a 2D frame structural modeling problem}
\label{Figure2}
\end{figure*}

The input to the multi-agent LLMs is a textual description of the 2D frame structural modeling problem. The description template, shown in \cref{Figure2}, includes four components: geometry, boundary conditions, load patterns, and material properties. The geometry section specifies the overall configuration, including the number of bays and the number of stories in each bay, as well as detailed dimensions such as bay spans and story heights. The boundary conditions section identifies the types and locations of supports. The load patterns section defines the load types, magnitudes, directions, and application locations. The material properties section provides the Young’s modulus, cross-sectional areas, and moments of inertia for columns and girders. Given this textural input, the multi-agent LLMs automatically generate the structural modeling scripts and invokes OpenSeesPy to perform the corresponding analysis. Performance evaluation focuses on three criteria: accuracy across ten repeated trails, efficiency of the inference process, and scalability to larger structural systems.

\subsection{Performance evaluation of sequential multi-agent architecture}

\begin{figure*}[htbp]
\centering
% \captionsetup{justification=centering}
\includegraphics[width=0.8\textwidth]{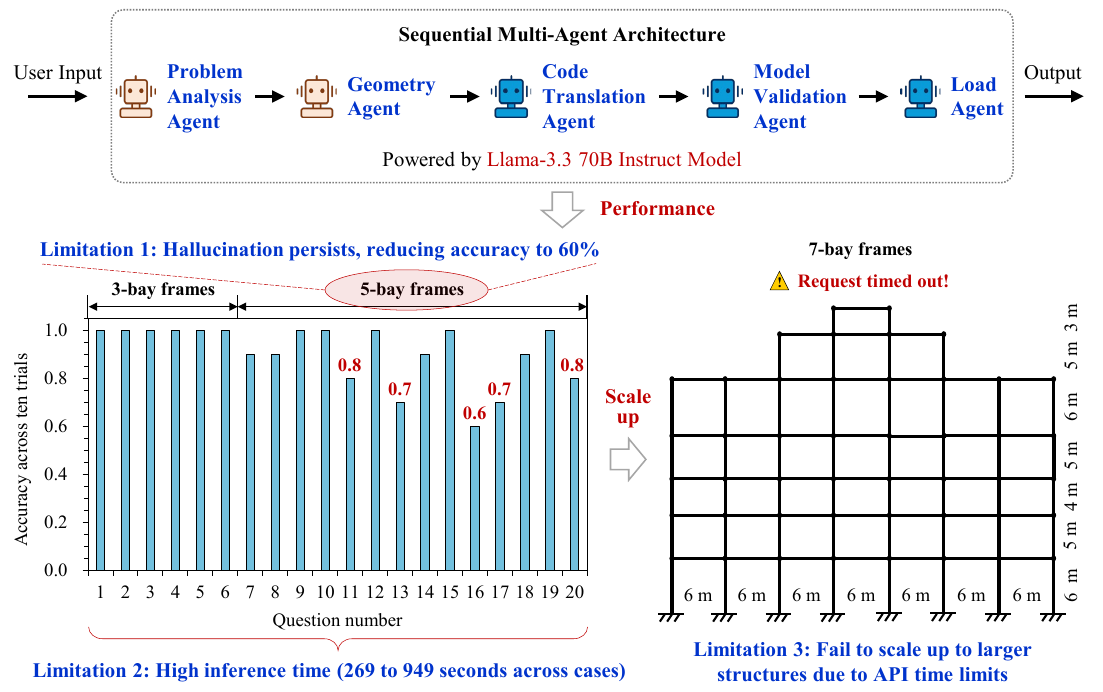}
\caption{Limitations of sequential multi-agent architecture in accuracy, efficiency, and scalability.}
\label{Figure3}
\end{figure*}

The multi-agent LLMs proposed in \cite{geng2025lightweight} adopt a sequential architectural design, in which the overall frame structural modeling task is decomposed into five subtasks: problem analysis, geometry assembly, code translation, model validation, and load application. Each subtask is handled by a specialized LLM agent, all powered by the Llama-3.3 70B Instruct model. Inter-agent communication follows a unidirectional pipeline: each agent receives the input from its predecessor and passes its output to the subsequent agent. The performance of this sequential architecture is evaluated using the benchmark dataset, where each problem is executed ten times to account for stochasticity of LLM outputs. The results show that the sequential multi-agent LLMs outperform leading general-purpose LLMs, such as Gemini-2.5 Pro and GPT-4o. However, notable limitations remain, particularly in terms of accuracy, efficiency, and scalability, as demonstrated in \cref{Figure3}.

First, while the sequential multi-agent architecture achieves error-free analysis for relatively simple structures such as frames with 3 bays, its performance deteriorates as structural complexity increases. For frames with 5 bays, the architecture exhibits unstable performance across cases, with accuracy dropping to as low as 60\%. This indicates that hallucination persists when the LLMs perform long-sequence inference. It is expected that as the number of stories and bays increases, the probability of hallucination will increase to an unacceptable level. Second, the sequential architecture incurs considerable inference time because each agent can proceed only after its predecessor has completed its operation. Despite offering improvements over manual coding, the benchmark problems still require 269-949 seconds to complete. This falls short of the efficiency expectations for real-world engineering workflows. Third, the architecture demonstrates limited scalability. When extended to larger structures, such as frames with seven bays and seven stories, the pipeline fails due to API time limits. Specifically, the geometry agent exceeds the 30-minute inference cap and triggers timeout errors. This failure illustrates the vulnerability of sequential architecture when the computational burden of a single agent escalates. Collectively, these limitations significantly hinder the practical applicability of the sequential multi-agent LLMs and underscore the need to improve accuracy, efficiency, and scalability in automated frame structural modeling.

\section{A Novel Multi-Agent Architecture for Frame Structural Modeling}
\label{sec:others}

\begin{figure*}[htbp]
\centering
% \captionsetup{justification=centering}
\includegraphics[width=0.8\textwidth]{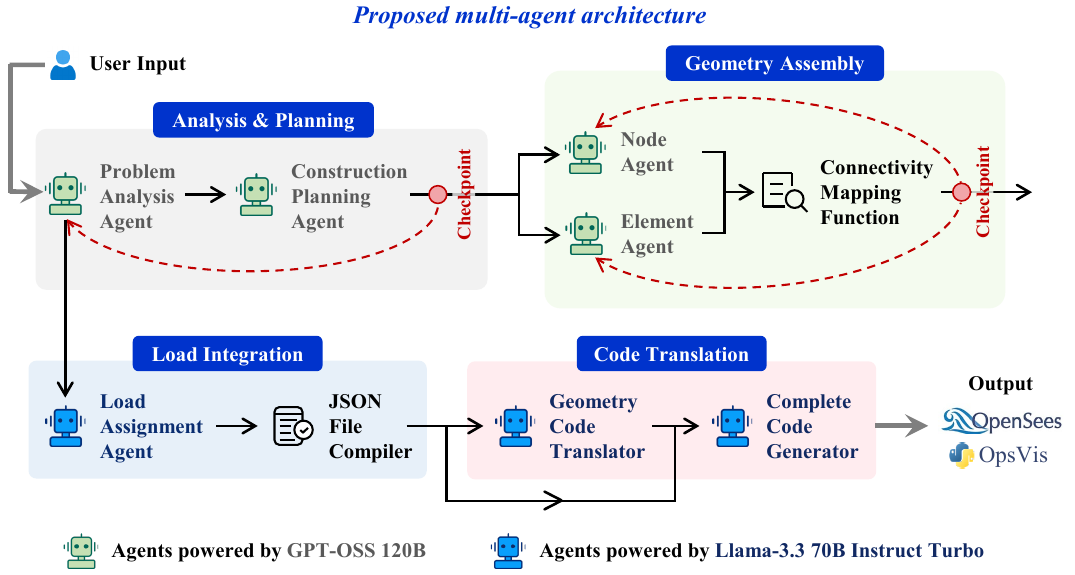}
\caption{A robust multi-agent architecture for LLMs to automate frame structural modeling.}
\label{Figure4}
\end{figure*}

To address the limitations of sequential architectural design, this section proposes a robust multi-agent architecture to automate frame structural modeling, as shown in \cref{Figure4}. The architecture receives a text description of the problem as input and processes it through four functional modules: analysis and planning, geometry assembly, load integration, and code translation. Each module includes one or more specialized LLM agents, whose roles are detailed in the following subsections. To improve robustness, checkpoints are embedded within the analysis and planning module and geometry assembly module. These checkpoints perform consistency checks and trigger regeneration when discrepancies are detected, thus mitigating error propagation to downstream agents. The architecture utilizes two lightweight LLM backbones: GPT-OSS 120B is assigned to agents that perform complex reasoning, whereas Llama-3.3 70B Instruct Turbo is used for tasks related to information translation and mapping. The rationale for this design choice is illustrated via a comparative experiment in Section 5. Following these modules, the multi-agent LLMs generate executable scripts, invoke OpenSeesPy \citep{zhu2018openseespy} for structural analysis, and utilize OpsVis \citep{kokot_opsvis_2024} to visualize the model geometry and structural responses.

\subsection{Analysis and planning}
\label{sec:overall_system}

The analysis and planning module includes two LLM agents: problem analysis agent and construction planning agent. The problem analysis agent extracts the key parameters from the user’s textual input and organizes them into a structured JSON file, as illustrated in Table~\ref{tab:json}. The JSON format comprises four components for downstream modeling: geometry, boundary conditions, material properties, and load patterns. Within the geometry section, the agent first identifies the total number of bays and stories in the frame, which will be used for subsequent consistency checks within the module. It then records detailed information such as bay index, span length, story count, and story heights. For boundary conditions, the agent classifies support types (e.g., pinned, roller, fixed) and specifies their locations. The material properties section captures five parameters: Young’s Modulus, cross-sectional areas, and moments of inertia for both columns and girders. The load section specifies the load type (point, distributed, or other), location, direction, and magnitude.

\begin{table}[htbp]
\centering
\captionsetup{skip=5pt}
\caption{JSON representations produced by the problem analysis agent, node agent, and element agent.}
\label{tab:json}
\begin{tabular}{p{4.4cm} p{5.4cm} p{5.4cm}}
\toprule
Problem analysis agent & Node agent & Element agent \\
\midrule
\ttfamily\scriptsize
\begin{minipage}[c]{\linewidth}
\begin{verbatim}
{
  "Geometry": {
    "Total_bays": <int>,
    "Total_stories": <int>,
    "Bay_data": [
      {
        "Bay": <int>,
        "Span": <float>,  
        "Story_count": <int>,
        "Heights": [<float>, ...] 
      }
      // Additional bays omitted
    ]
  },
  "Supports": {
    "Type": "<string>",   
    "Location": "<string>"   
  },
  "Material": {
    "E": <float>,        
    "A_col": <float>,     
    "A_gir": <float>,        
    "I_col": <float>,     
    "I_gir": <float>           
  },
  "Loads": {
    "Type": "<string>",           
    "Location": "<string>",    
    "Direction": "<string>",        
    "Magnitude": <float>      
  }
}

\end{verbatim}
\end{minipage}
&
\ttfamily\scriptsize
\begin{minipage}[c]{\linewidth}
\begin{verbatim}
{
  "Construction_steps": [
    {
      "Step_number": <int>,
      "Bay_number": <int>,
      "Story_number": <int>,
      "Step_type": "<string>",
      "Nodes": [
        {
          "ID": <int>, 
          "x": <float>, 
          "y": <float>, 
          "Description": "<string>"
        }
        // Additional nodes omitted
      ],
      "Boundary_conditions": [
        {
          "Node_ID": <int>, 
          "Constraints": "<string>"
        }
        // Additional constraints omitted
      ]
    }
    // Additional steps omitted
  ]
}
\end{verbatim}
\end{minipage}
&
\ttfamily\scriptsize
\begin{minipage}[c]{\linewidth}
\begin{verbatim}
{
  "Construction_steps": [
    {
      "Step_number": <int>,.      
      "Bay_number": <int>,
      "Story_number": <int>,
      "Step_type": "<string>",
      "Elements": [
        {
          "ID": <int>,
          "Coord_i": [[<float>, <float>], 
          "Coord_j": [<float>, <float>]],
          "Description": "<string>"
        }
        // Additional elements omitted
      ]
    }
    // Additional steps omitted
  ]
}
\end{verbatim}
\end{minipage}
\\
\bottomrule
\end{tabular}
\end{table}

\begin{figure*}[htbp]
\centering
% \captionsetup{justification=centering}
\includegraphics[width=0.8\textwidth]{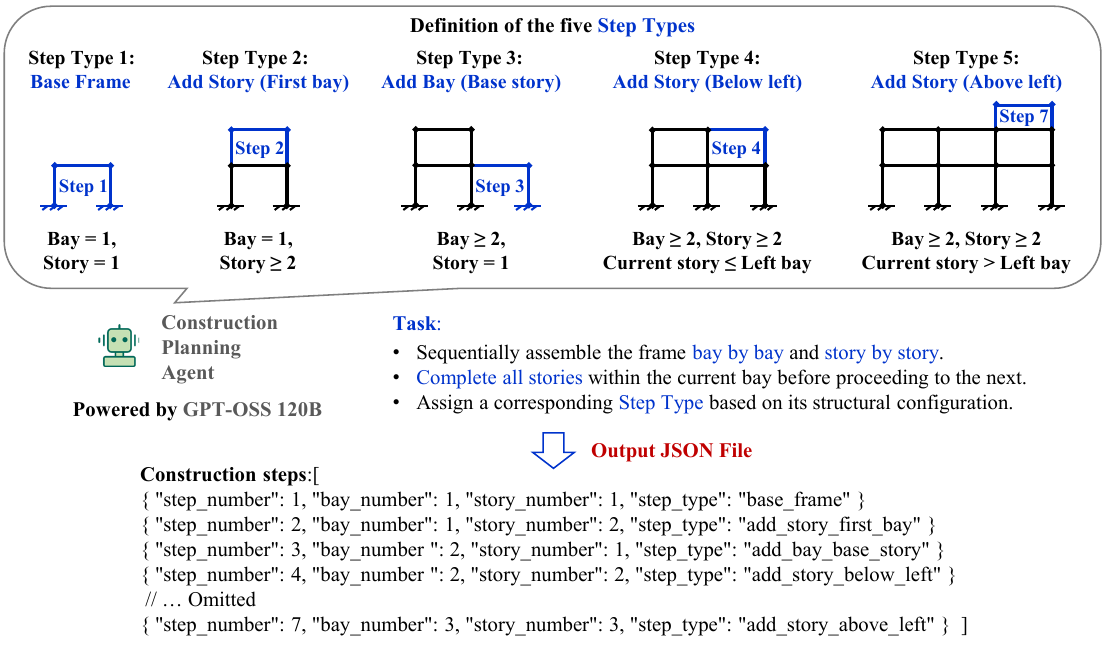}
\caption{Workflow of construction planning agent for generating a stepwise assembly plan.}
\label{Figure5}
\end{figure*}

The construction planning agent receives the output from the problem analysis agent and generates a stepwise plan for assembling the 2D frame, as demonstrated in \cref{Figure5}. The planning process follows a bay-by-bay, story-by-story logic: the agent constructs all stories in the current bay before proceeding to the next, and within each bay, stories are constructed from bottom to top. For each construction step, the agent assigns a step type that instructs downstream agents to apply the appropriate rule. These step types are derived from expert domain knowledge and capture five possible conditions encountered in 2D frame assembly. Specifically, step type 1 is assigned when constructing the first story of the first bay. Step type 2 applies to additional stories \( (\textnormal{story} \ge 2) \) within the first bay. Step type 3 is used when expanding the first story in subsequent bays \( (\textnormal{bay} \ge 2) \). For higher stories in subsequent bays, step type 4 is assigned if the height of the current story is less than or equal to that of the adjacent left bay, while step type 5 is used when it exceeds the height of the left bay.

The output of the construction planning agent is a JSON file containing a sequence of construction steps, each with an assigned step number, bay number, story number and step type. To ensure the consistency of this module’s output, a checkpoint is placed following the planning stage, as shown in \cref{Figure4}. This checkpoint performs two consistency checks: (a) whether the maximum bay number in the generated plan matches the total number of bays in the problem analysis, and (b) whether the total number of construction steps equals the total number of stories in the problem analysis. If either check fails, the pipeline re-executes both the problem analysis and construction planning procedures, with up to five retries permitted. Because these two agents involve complex semantic reasoning, both are powered by the GPT-OSS 120B model.

\subsection{Geometry assembly}
\label{sec:overall_system}

The geometry assembly module is designed to determine the nodes and elements required to construct the 2D frame structure. As shown in \cref{Figure4}, this module consists of two LLM agents, the node agent and the element agent, as well as a connectivity mapping function. To improve computational efficiency, the two agents operate in parallel and receive same input from two upstream agents: (a) problem analysis agent, which provides geometric parameters such as bay spans and story heights and boundary conditions, and (b) construction planning agent, which specifies the assembly step and associated step types. Specifically, the node agent is tasked with deriving node identifiers, spatial coordinates, and boundary conditions, while the element agent determines element identifiers, types, and end coordinates. 

\begin{figure*}[htbp]
\centering
% \captionsetup{justification=centering}
\includegraphics[width=0.8\textwidth]{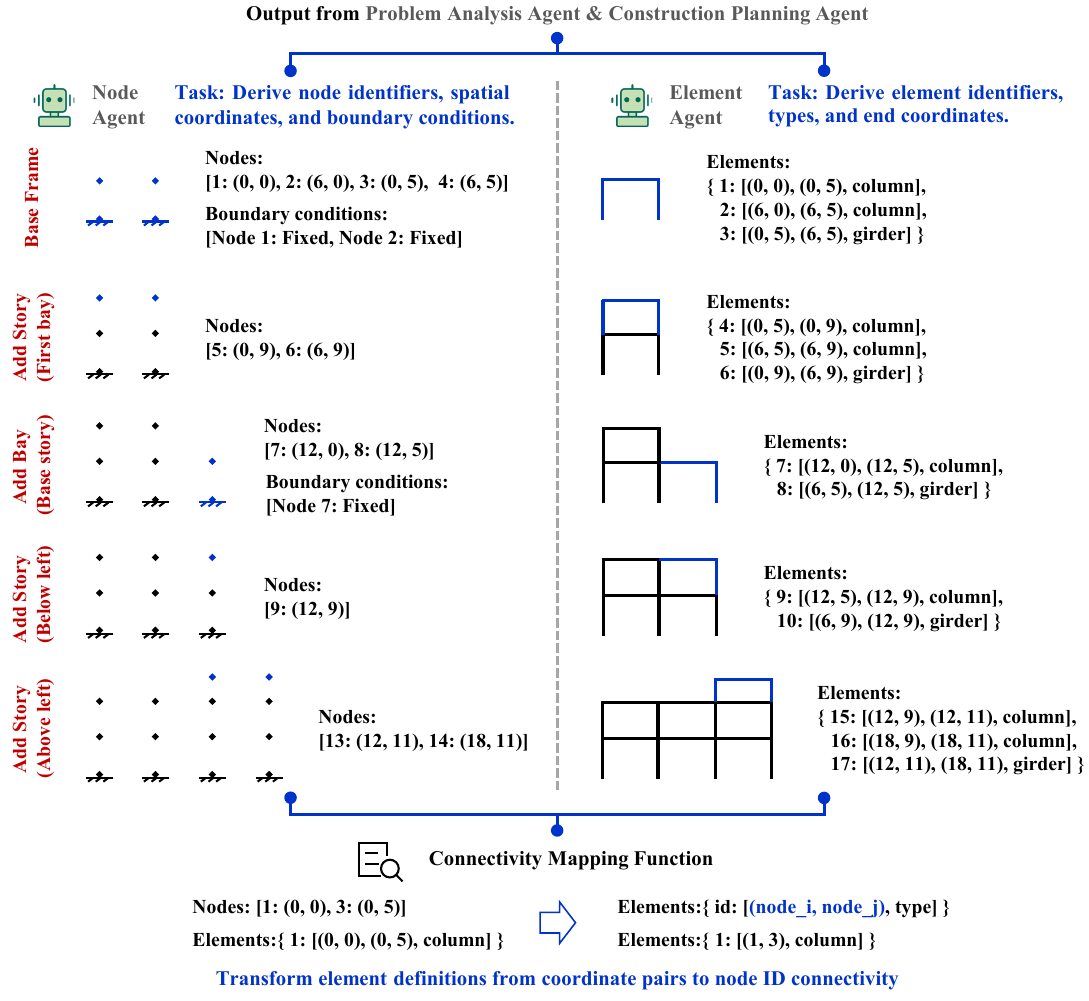}
\caption{Workflow of the geometry assembly module, in which the node and element agents operate in parallel, followed by a connectivity mapping function.}
\label{Figure6}
\end{figure*}

Both agents proceed step-by-step according to the construction plan, and their actions within each step are instructed by the step type. For each step type, a dedicated set of rules is defined by domain experts to separately guide the node agent and the element agent. These rules prescribe the number of nodes or elements to be added and provide formulas to calculate node coordinates from the input dimensions. For instance, when constructing the first story of the first bay, the step type is identified as “base frame”. The node agent applies the corresponding rule to define four nodes and compute their coordinates using the given span and height. It then identifies the nodes with zero y‑coordinate as base nodes and assigns fixed supports. In parallel, the element agent applies its rule to generate two columns and one girder and determine their end coordinates. \cref{Figure6} provides illustrative examples for each rule set. 

Following the parallel generation of nodes and elements, the module performs post-processing and validation steps. First, a connectivity mapping function is executed to transform the representation of each element into a format compatible with the OpenSeesPy syntax. Specifically, this deterministic Python function maps endpoint coordinate pairs (e.g., $[(x_1, y_1), (x_2, y_2)]$) to node ID connectivity (e.g., $[\mathrm{node}_i, \mathrm{node}_j]$). Then, a checkpoint is applied to ensure geometric consistency. The checks include (a) duplicate nodes, (b) duplicate elements, (c) element coordinates with no matching node, and (d) nodes that are not referenced by any element. These validations are also implemented using Python functions to enhance robustness. If any discrepancy is detected, the pipeline re-invokes both the node and element agents, with a maximum of five regeneration attempts. A template for the structured JSON outputs of the node and element agents is provided in Table~\ref{tab:json}. Because the node and element agents perform rule-based sequential reasoning, GPT-OSS 120B is selected as the backbone LLM.

\subsection{Load integration}
\label{sec:overall_system}

The load integration module includes two components: the load assignment agent and the JSON file compiler, as illustrated in \cref{Figure4}. The load assignment agent is responsible for translating abstract load descriptions into a structured format that conforms to the load application syntax in OpenSeesPy. It receives input from two upstream sources: (a) problem analysis agent, which provides load attributes, including type, location, direction, and magnitude, and (b) geometry assembly module, which records the definitions of nodes and elements, including their identifiers, coordinates, and types. Using these inputs, the agent parses the load descriptions and determines the corresponding structural components for load application, such as assigning point loads to nodes and distributed loads to elements. Given that this task involves consistent information mapping, the Llama-3.3 70B Instruct Turbo model is selected as the agent’s backbone due to its strong performance in instruction-following tasks.

\begin{figure*}[htbp]
\centering
% \captionsetup{justification=centering}
\includegraphics[width=0.8\textwidth]{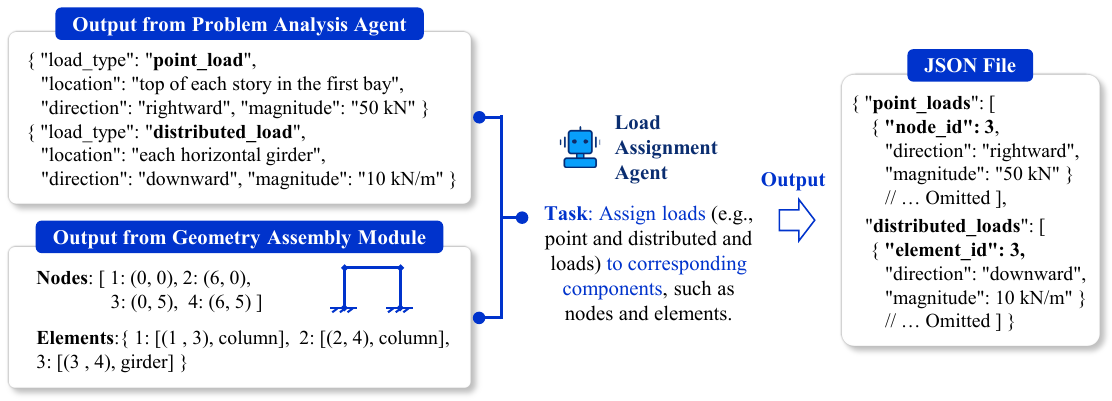}
\caption{Workflow of the load assignment agent for mapping loads to corresponding structural components.}
\label{Figure7}
\end{figure*}

The output of the load assignment agent is a structured JSON file that links each load, characterized by type, magnitude, and direction, to the corresponding structural components, as demonstrated in \cref{Figure7}. While the benchmark dataset includes a representative load pattern, the agent is designed to handle diverse loading cases. More examples are presented in Section 4.3. Following load assignment, the JSON file compiler integrates the load data with other structured JSON outputs from upstream modules, such as material properties from the problem analysis agent and node and element definitions from the geometry assembly module. This is implemented via a deterministic Python utility. The compiled file encapsulates all information required for frame modeling in OpenSeesPy and serves as inputs to the code translation module for generating the executable scripts.

\subsection{Code translation}
\label{sec:geometry_agent}

\begin{figure*}[htbp]
\centering
% \captionsetup{justification=centering}
\includegraphics[width=0.8\textwidth]{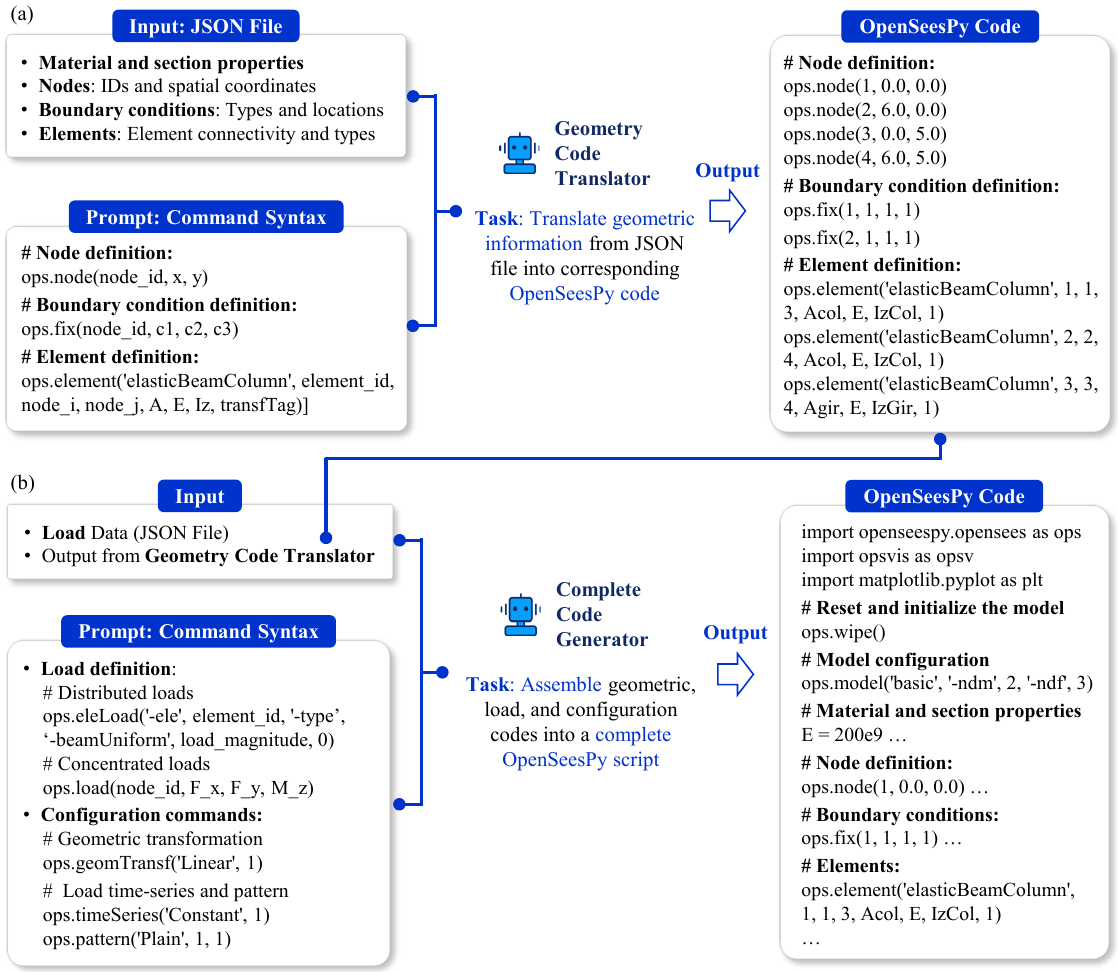}
\caption{Workflow of the code generation module, in which the geometry code translator converts geometric information into OpenSeesPy code and complete code generator integrates the geometric code with load and configuration commands to produce executable OpenSeesPy scripts.}
\label{Figure8}
\end{figure*}

The code translation module is tasked with converting the structured JSON file into a complete, executable OpenSeesPy script. To mitigate the risk of hallucinations associated with long-context reasoning, the task is decomposed into two subtasks and assigned to two specialized LLM agents: the geometry code translator and the complete code generator, as shown in \cref{Figure4}. The geometry code translator focuses on transforming geometric information, including nodes, boundary conditions, and elements, into corresponding OpenSeesPy code. This process is guided by a prompt that specifies the required command syntax and parameter specifications. The outputs of the translator include three code blocks, as illustrated in \cref{Figure8} (a).

Following this, the complete code generator assembles the OpenSeesPy script by integrating the translated geometric code with additional blocks for load definitions and model configuration. Its role lies in twofold. First, it processes load data from the JSON file to generate point and distributed load commands using OpenSeesPy syntax. Then, it integrates the geometry and load code with mandatory configuration commands, such as geometric transformation and time series and patterns of loads. The resulting output is a complete and executable OpenSeesPy script, as shown in \cref{Figure8} (b). Given that both agents perform deterministic translation tasks, the Llama-3.3 70B Instruct Turbo model is used as the backbone LLM for its strong adherence to syntax rules.

\section{Results and Discussion}

\subsection{Performance of the proposed multi-agent architecture}

\begin{figure*}[htbp]
\centering
% \captionsetup{justification=centering}
\includegraphics[width=0.8\textwidth]{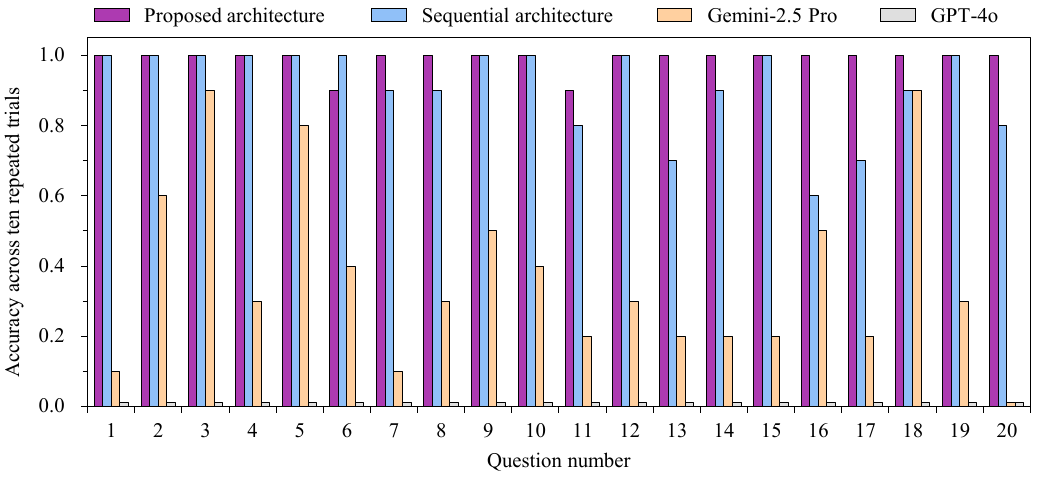}
\caption{Performance comparison of the proposed multi‑agent architecture with the sequential architecture and two state‑of‑the‑art general-purpose LLMs.}
\label{Figure9}
\end{figure*}

The performance of the proposed multi-agent architecture is evaluated using the benchmark dataset. As shown in \cref{Figure9} and Table~\ref{tab:avg_accuracy}, the architecture exhibits strong reliability and robustness across 20 frame analysis problems. It achieves error-free predictions for 18 cases and 90\% accuracy for the remaining two. These results indicate that the architecture effectively coordinates specialized agents and substantially reduces the risk of hallucinations. In comparison, the sequential multi-agent architecture attains 80\% accuracy for most cases, but its performance drops to 60\% for the frame configuration of 2-3-1-4-5. This highlights its deficiencies in robustness under higher structural complexity. Additionally, the proposed architecture significantly outperforms two state-of-the-art general-purpose LLMs, including Gemini 2.5 Pro and GPT 4o. These two LLMs receive the textual descriptions of problems and are prompted to generate executable OpenSeesPy code. It shows that the general-purpose LLMs struggle to perform domain-specific structural modeling tasks. Gemini achieves an average accuracy of only 37\%, while GPT fails to produce the correct code across all test cases. 

\begin{table}[htbp]
\centering
\captionsetup{skip=5pt}
\caption{Average accuracy across 20 benchmark problems for different architectures and backbone LLMs.}
\label{tab:avg_accuracy}
\begin{tabular}{
m{2cm}
>{\centering\arraybackslash}m{1.7cm}
>{\centering\arraybackslash}m{1.7cm}
>{\centering\arraybackslash}m{1.4cm}
>{\centering\arraybackslash}m{1.1cm}
>{\centering\arraybackslash}m{1.3cm}
>{\centering\arraybackslash}m{0.9cm}
>{\centering\arraybackslash}m{1.0cm}
>{\centering\arraybackslash}m{1.0cm}
}

\toprule
& \multicolumn{4}{c}{Architecture comparison} 
& \multicolumn{4}{c}{Backbone LLM comparison} \\
\cmidrule(l{3pt}r{3pt}){2-5} 
\cmidrule(l{3pt}r{3pt}){6-9}

Metric 
& Proposed\newline architecture 
& Sequential\newline architecture 
& Gemini-\newline 2.5 Pro 
& GPT-4o 
& GPT \&\newline Llama 
& GPT 
& Llama 
& Qwen \\

\midrule
Avg.\ accuracy & \textbf{99\%} & 91\% & 37\% & 0\% & \textbf{99\%} & 90\% & 79\% & 69\% \\
\bottomrule
\end{tabular}
\end{table}

\begin{figure*}[t]
\centering
% \captionsetup{justification=centering}
\includegraphics[width=0.8\textwidth]{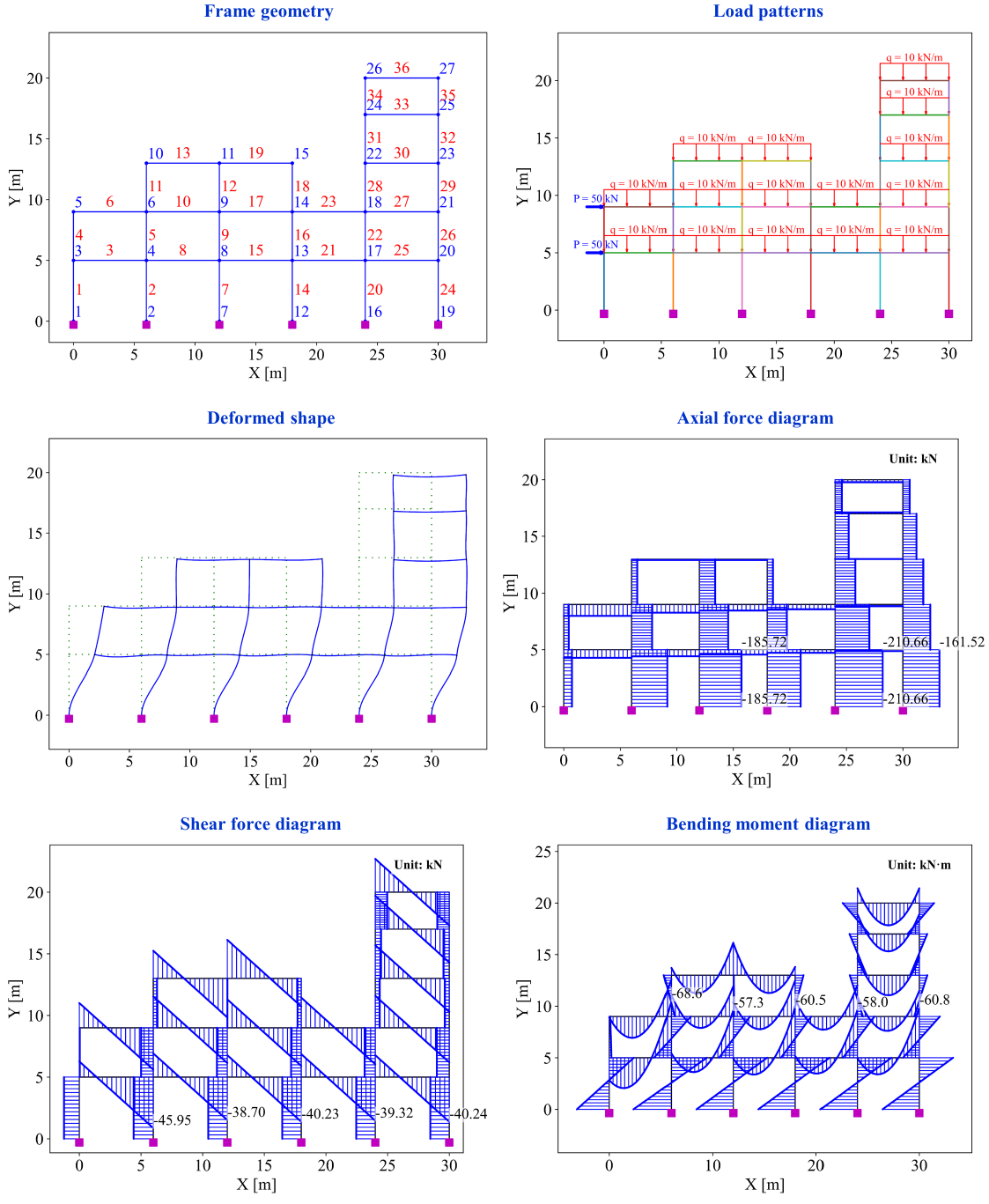}
\caption{Visualizations of the proposed multi-agent LLMs, including geometry, loading, deformation, and internal force diagrams.}
\label{Figure10}
\end{figure*}

The proposed multi-agent LLMs automatically execute the generated OpenSeesPy scripts and visualize the analysis results using OpsVis. As illustrated in \cref{Figure10}, a representative frame configuration of 2-3-1-4-5 is used to showcase six visual outputs: frame geometry, load patterns, deformed shape, axial force diagram, shear force diagram, and bending moment diagram. Specifically, the geometry and load visualizations provide a clear representation of nodes, elements, boundary conditions, and applied loads. They enable engineers to identify potential modeling errors and implement corrective actions accordingly. The deformed shape is computed from nodal displacements and scaled for intuitive interpretation. These nodal displacements (i.e., elastic deflections) are also recorded in the output files. The visualizations enhance the usability of the multi-agent LLMs by providing an integrated workflow from model generation to analysis and interpretation, making it well-suited for both engineering practice and educational applications.

\subsection{Effects of backbone LLMs on performance}

The proposed multi-agent architecture utilizes two backbone LLMs to enhance task alignment. Specifically, GPT-OSS 120B is assigned to agents that perform complex reasoning, including problem analysis agent, construction planning agent, node agent, and element agent. In contrast, Llama-3.3 70B Instruct Turbo powers agents tasked with information mapping and translation, including load assignment agent, geometry code translator, and complete code generator. This subsection aims to justify the rationale for this selection by benchmarking its performance against three alternatives. Each alternative is powered by a single LLM: GPT-OSS 120B, Llama-3.3 70B Instruct Turbo, and Qwen-3 235B Instruct. All systems are evaluated using the benchmark dataset, with each problem tested for ten repeated trials.

\begin{figure*}[htbp]
\centering
% \captionsetup{justification=centering}
\includegraphics[width=0.8\textwidth]{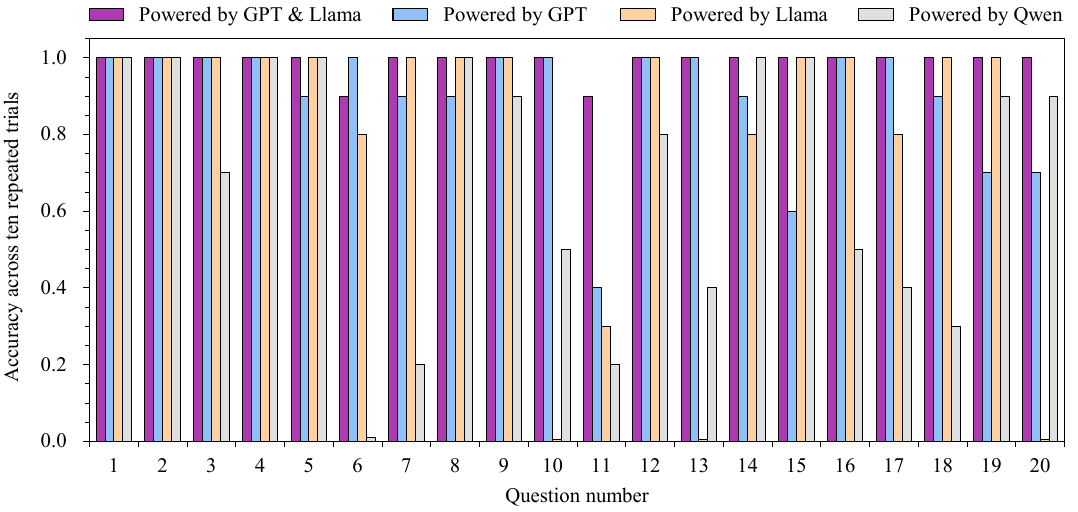}
\caption{Performance comparison of the multi-agent architecture powered by various LLMs.}
\label{Figure11}
\end{figure*}

\cref{Figure11} and Table~\ref{tab:avg_accuracy} present the performance comparison between the proposed multi-agent LLMs and three alternatives across 20 test cases. The results indicate that the proposed architecture consistently achieves near-perfect accuracy, demonstrating strong generalization and robustness. In contrast, alternatives powered by a single LLM exhibit significant performance variability across different frame configurations. Among the three alternatives, the GPT-powered architecture achieves the highest average accuracy of 90\%. However, its performance is unstable, dropping to 40\% for the 3‑5‑2‑3‑5 frame and 60\% for the 5‑3‑2‑4‑1 frame. The Llama-powered architecture yields an average accuracy of 79\% but fails entirely on frames with particular spatial geometry such as 1-2-3-1-5, 2-4-3-2-5, and 2-4-3-5-1. The Qwen-powered architecture exhibits the lowest average accuracy of 69\%. While it performs reliably on frames with 3 bays, the performance degrades significantly on frames with 5 bays, where accuracy drops below 50\% in half of the cases. These findings show that the proposed prompt template is effective and interpretable across general-purpose LLMs because all three alternatives exhibit high accuracy in frames with 3 bays and certain cases with 5 bays such as 2-2-3-1-2 and 3-2-3-2-3. However, the varied failure patterns observed across three alternatives underscore the sensitivity of LLMs to prompt design and problem configuration. This reveals a critical vulnerability of multi-agent architectures powered by a single LLM.

To uncover the causes of performance degradation in alternatives powered by single LLMs, the intermediate outputs of all agents are extracted for detailed error analysis. The results reveal diverse hallucination patterns across LLMs, as illustrated in the Appendix. Specifically, the GPT-powered architecture exhibits hallucinations primarily in mapping and translation tasks. In case 1, the element agent correctly defines geometric attributes in the JSON file, but the geometry code translator introduces a duplicated element when generating OpenSeesPy code, resulting in execution failures. In case 2, the load assignment agent fails to map the distributed load to element 25, despite the element being correctly defined as a girder. In contrast, hallucinations in Llama-powered architecture concentrate on complex reasoning tasks. In case 1, the construction planning agent redundantly produces steps that had already been included in prior planning. This leads to a mismatch between the total number of stories and construction steps, ultimately causing script failure. In case 2, the node agent loses mathematical precision during long-context reasoning. Although the story heights are correctly specified as 5, 4, 3, 2, and 1 meters, the agent assigns incorrect cumulative elevations of 13, 16, and 18 meters for the third, fourth, and fifth floors, respectively.

The Qwen-powered architecture exhibits broad vulnerability, producing hallucinations in both reasoning and translation tasks. In case 1, the construction planning agent fails to maintain logical consistency during the long-sequence reasoning task. It generates repeated steps during the construction of the first story in bays 3, 4, and 5, and misclassifies the step type for bay 2. In case 2, during the code translation process, the geometry code translator resets a valid horizontal coordinate to zero, resulting in a duplicate node and an incorrect diagonal connection in the frame. Collectively, these analyses confirm that no single LLM exhibits consistent reliability across all subtasks in structural analysis. Each model demonstrates task-specific strengths: GPT excels in complex planning and reasoning, whereas Llama performs reliably in information mapping and translation. To harness these complementary strengths, the proposed multi-agent architecture allocates backbone LLMs to agents according to their specific task type. This design mitigates the limitations of single-LLM systems and enhances robustness across structural configurations.

\subsection{Adaptation to diverse linguistic styles}

\begin{figure*}[t!]
\centering
% \captionsetup{justification=centering}
\includegraphics[width=0.79\textwidth]{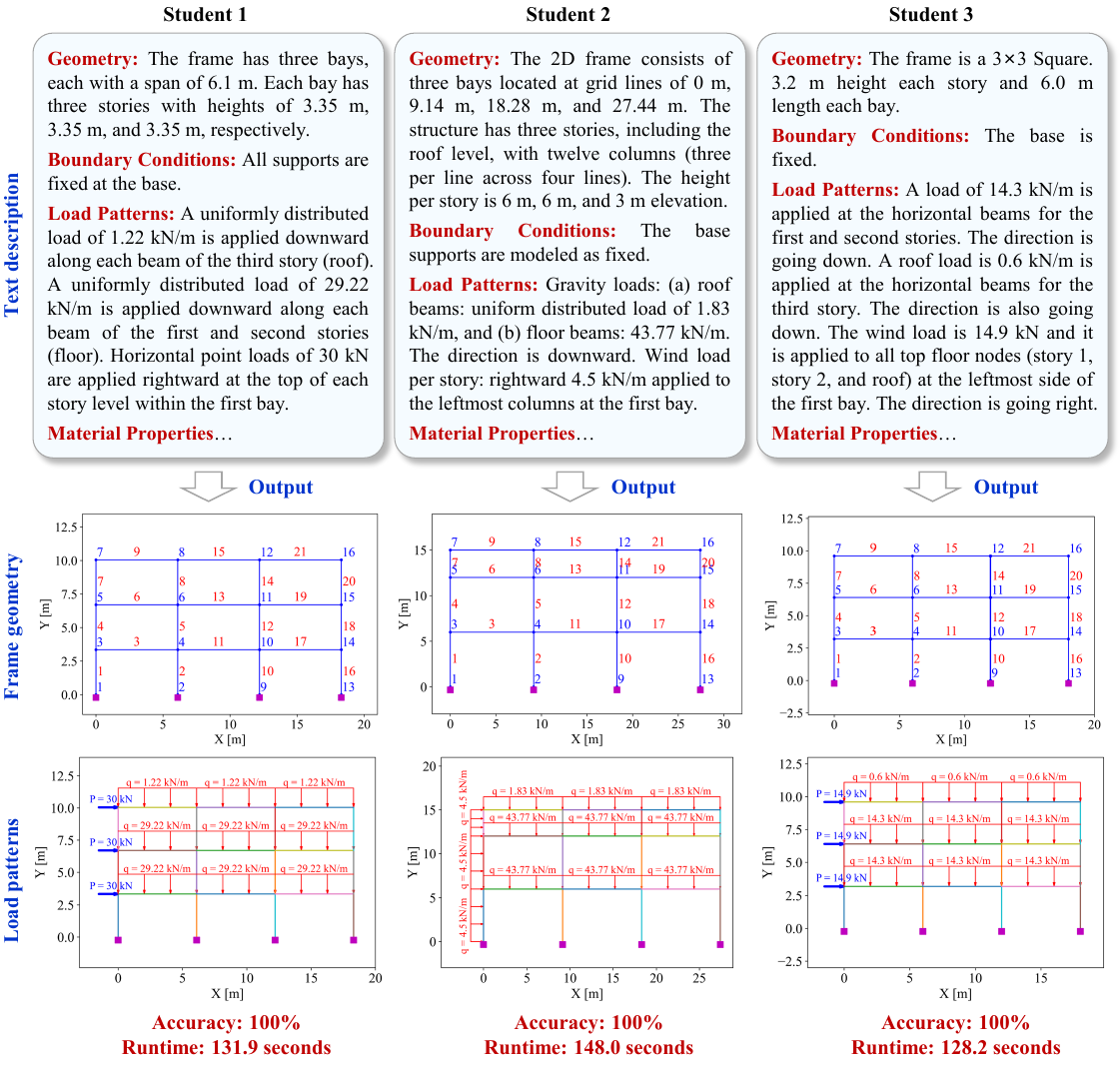}
\caption{Adaptation of the proposed architecture to diverse input styles provided by three students.}
\label{Figure12}
\end{figure*}

The proposed multi-agent architecture provides an end-to-end workflow that receives natural language input and outputs structural analysis results. This functionality significantly lowers the entry barrier for non-expert users by enabling task specification via flexible textual descriptions. However, this flexibility raises a key question: can the system maintain robust and reliable performance across diverse linguistic styles? To evaluate this capability, a pilot test is conducted with three students from the School of Architecture at the University of Miami, as shown in \cref{Figure12}. Unlike civil engineering students, these participants possess only introductory-level knowledge of structural mechanics, making them well-suited for assessing the usability from a non-specialist perspective.

Specifically, student 1 follows the provided input template to describe the geometry, boundary conditions, and load patterns. Notably, the student assigns distinct distributed loads across floors: 1.22 kN/m for the third floor and 29.22 kN/m for the first and second floors. Additionally, horizontal point loads of 30 kN are applied to the top floor nodes in the first bay. Student 2 describes the structural geometry using gridlines, which is a conventional practice in architectural and structural design. Instead of explicitly specifying bay spans, this student describes the locations of vertical gridlines and indicates the number of columns per line to imply the number of stories. The load patterns include gravity loads of 1.83 kN/m on the first and second floors and 43.77 kN/m on the third floor. Wind loads are represented as distributed loads of 4.5 kN/m applied to the leftmost columns, which differs from the point loads used by student 1. Student 3 provides a concise geometric description, referring to the frame as a 3×3 square with equal bay span and story height. The load pattern matches that of student 1 in type but differs in magnitude: a distributed load of 0.6 kN/m on the third floor, 14.3 kN/m on the first and second floors, and 14.9 kN point loads on the façade.

Despite substantial variations in problem descriptions provided by users, the proposed architecture consistently generates correct structural analysis results across all ten repeated trials for each case, as illustrated in \cref{Figure12}. This robust performance demonstrates the effectiveness of the problem analysis agent in performing semantic reasoning on diverse user inputs. The agent can extract key parameters required for structural modeling and organize them into a standardized JSON format. This structured representation ensures that downstream agents operate reliably, regardless of differences in user’s language styles or levels of domain expertise. These results indicate the strong adaptability and robustness of the proposed architecture, highlighting its huge potential for broader adoption beyond engineering professionals.

\subsection{Scalability to larger structural systems} 
In addition to accuracy, another advantage of the proposed multi-agent architecture over sequential design is its scalability to larger structural systems. As mentioned in Section 2, the sequential pipeline often encounters timeout errors when handling large-scale structures due to prolonged reasoning times that exceed API limits. These errors are commonly triggered by the geometry agent, which is tasked with defining node coordinates and element connectivity simultaneously. To overcome this limitation, the proposed architecture decomposes the geometry assembly task into two independent sub-tasks, each assigned to a dedicated agent: the node agent and the element agent. This task decomposition significantly enhances the overall efficiency, improving the capacity to scale up.

\begin{figure*}[htbp]
\centering
% \captionsetup{justification=centering}
\includegraphics[width=0.8\textwidth]{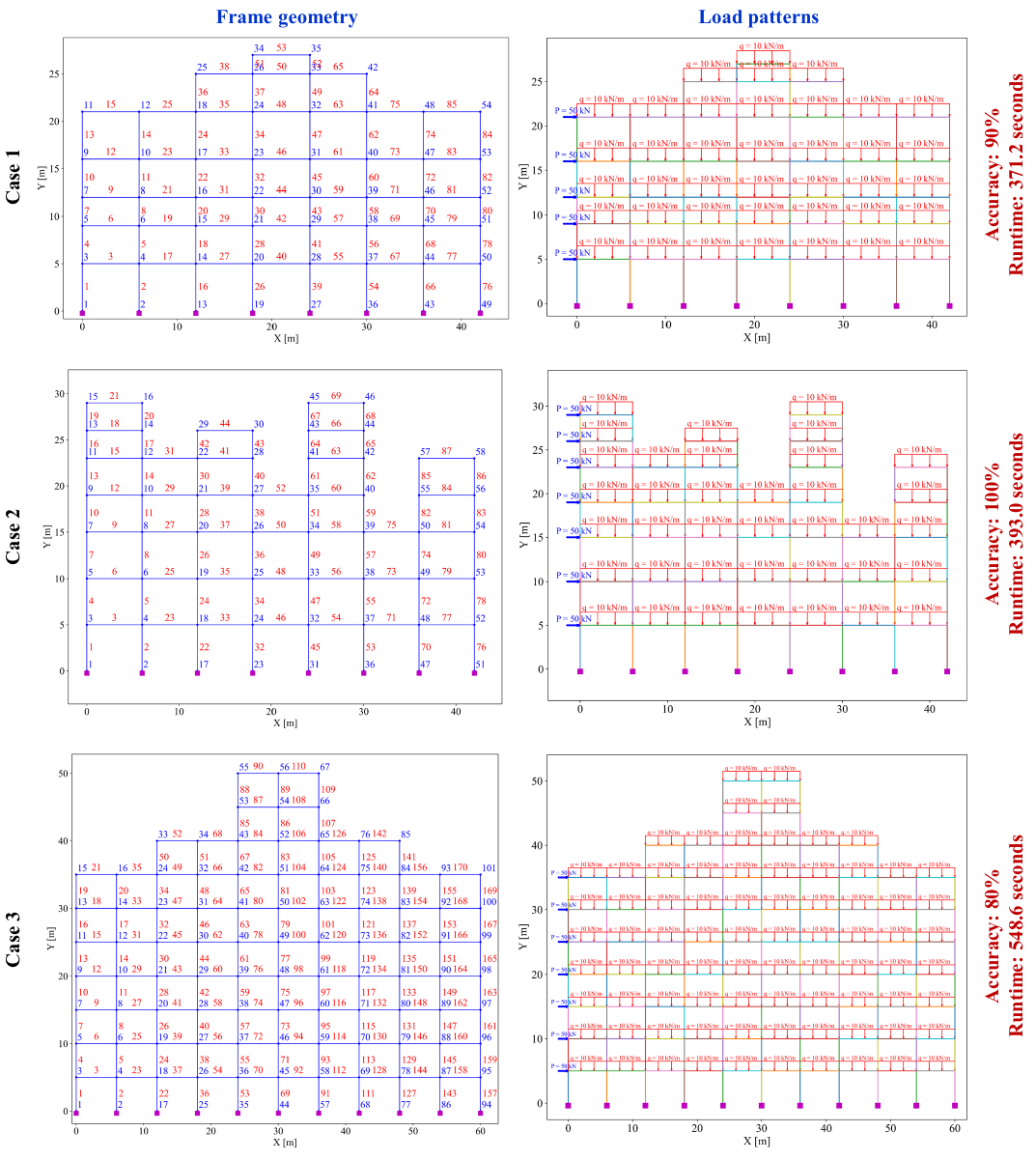}
\caption{Scalability of the proposed architecture across three representative structural configurations.}
\label{Figure13}
\end{figure*}

\cref{Figure13} illustrates the scalability of the proposed architecture through three representative cases. The configurations of these cases are as follows. Case 1 features a frame with 7 bays and a symmetric profile, comprising 5, 5, 6, 7, 6, 5, and 5 stories from left to right. Case 2 introduces random asymmetry in a frame with 7, 5, 6, 4, 7, 3, and 5 stories. Case 3 further scales up to a ten-bay, ten-story frame with story counts of 7, 7, 8, 8, 10, 10, 8, 8, 7, and 7. All three cases adopt a consistent load pattern from the benchmark dataset: each girder is subjected to a uniformly distributed downward load of 10 kN/m, while each top floor node at the leftmost bay is subjected to a rightward point load of 50 kN. The results demonstrate that the proposed architecture maintains strong performance under increasing structural complexity, achieving accuracy rates of 90\%, 100\%, and 80\% across the three cases, respectively. These findings affirm the effectiveness of the task decomposition strategy in mitigating computational bottlenecks while ensuring robustness and scalability for large-scale structural systems.

\subsection{Runtime and cost}

\begin{figure*}[htbp]
\centering
% \captionsetup{justification=centering}
\includegraphics[width=0.45\textwidth]{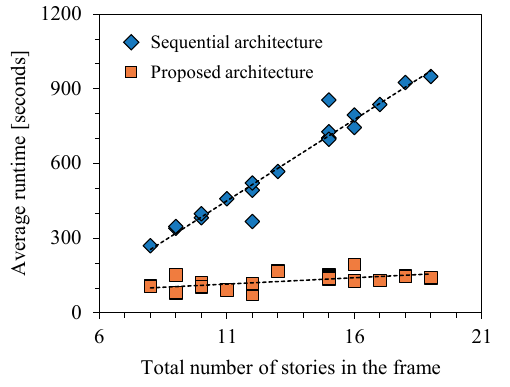}
\caption{Runtime comparison between the proposed multi-agent architecture and the sequential design.}
\label{Figure14}
\end{figure*}

Beyond accuracy and scalability, computational efficiency is another critical factor for evaluating the practicality of automated structural modeling and analysis. To this end, the proposed multi-agent architecture is compared with sequential design in terms of runtime across 20 benchmark problems. For each case, the average runtime is computed over ten repeated trials. \cref{Figure14} clearly demonstrates the substantial improvement in computational efficiency provided by the proposed architecture over the sequential design. Specifically, the sequential architecture exhibits runtime ranging from 269.2 to 949.0 seconds, whereas the proposed architecture reduces this range to between 75.4 and 194.2 seconds. A particularly illustrative example is the 3-4-5-4-3 frame, where the sequential architecture requires 949.0 seconds, while the proposed architecture completes the task in just 140.9 seconds, which is an 85\% reduction in inference time. Additionally, the proposed architecture maintains efficient computational performance when scaling to larger structural systems. As illustrated in \cref{Figure13}, the total inference times for three large-scale cases are 371.2, 393.0, and 548.6 seconds, respectively. These results indicate that the proposed architecture exhibits an approximately linear growth in runtime with increasing structural complexity, which is a highly desirable property for practical applications in engineering workflows.

The economic cost of the proposed architecture is further evaluated through token consumption analysis using the benchmark dataset. For each of the 20 problems, total input and output token usage is recorded for LLMs: GPT-OSS 120B and Llama-3.3 70B Instruct Turbo. Specifically, the 3-2-3 frame incurs the lowest token usage. GPT consumes 6,867 input tokens and 3,394 output tokens, while Llama uses 8,323 input tokens and 3,325 output tokens. In contrast, the 3-4-5-4-3 frame yields the highest token consumption. GPT processes 9,645 input tokens and 8,333 output tokens, whereas Llama consumes 13,410 input tokens and 5,778 output tokens. Based on the price information provided by Together AI (in 2026 US\$), GPT costs \$0.15 per million input tokens and \$0.60 per million output tokens, while Llama is priced at \$0.88 per million tokens for both input and output. Accordingly, the total cost per benchmark problem ranges from \$0.013 to \$0.023 (2026 US\$). These results highlight the economic viability of the proposed architecture, demonstrating that accurate structural modeling and analysis can be achieved at minimal cost using lightweight LLMs.

\section{Limitations and Future work}

Despite the reliable, efficient, and robust performance of the proposed multi-agent LLMs for automated structural analysis of frames, several limitations remain. First, although the proposed prompt templates are effective for the selected backbone LLMs, they fail to maintain consistent reliability across other models, as discussed in Section 4.2. This shows the sensitivity of LLMs to prompt engineering, which is a widely recognized challenge in existing literature \citep{sclar2023quantifying, chatterjee2024posix}. Therefore, future work should focus on developing an automated prompt optimization framework that can adapt domain-specific instructions to diverse LLMs. Second, the current architecture is restricted to the structural analysis of rectangular frames with vertical columns and horizontal beams. It lacks the ability to model structural components such as diagonal bracing and cantilevers, which are commonly used in real-world practice. Future research should expand the architecture’s capabilities to handle complex structural typologies to improve its generalizability and practical utility. Third, the current structural analysis scope is limited to linear elastic behavior under static loading conditions. While adequate for preliminary design and verification tasks, it does not address dynamic effects (e.g., wind and seismic loads) or nonlinear analysis. Incorporating these capabilities is essential for broader applicability of the LLM in advanced structural design workflows.

\section{Summary and Conclusions}

This paper proposes a novel large language model (LLM)-based multi-agent architecture for automated structural analysis of 2D frames. The architecture is designed to address three limitations of existing LLMs: (a) unstable performance and hallucinations across diverse structural configurations, (b) low computational efficiency, and (c) limited scalability to larger structural systems. The proposed multi-agent architecture is as follows. The problem analysis agent interprets the user’s description and extracts key parameters required for structural analysis. Then, the construction planning agent formulates a stepwise plan for assembling the frame. Next, the node agent and element agent operate in parallel to define node coordinates and element connectivity. This is followed by a load assignment agent, which applies nodal and elemental loads and compiles all material, geometric, boundary, and load information into a JSON file. This JSON file is then processed by two translation agents to produce the geometry code and the complete OpenSeesPy scripts. To improve robustness, checkpoints are embedded after the planning stage and the node/element generation stage. When inconsistencies are detected, the architecture returns to the previous step and regenerates the outputs. The performance of the proposed architecture is assessed using a benchmark dataset of 20 representative frame structural analysis problems. The key findings are summarized below:

\begin{itemize}
    \item The proposed multi-agent architecture demonstrates reliable and robust performance on the benchmark dataset, achieving 100\% accuracy in 18 cases and 90\% in the remaining two. These results not only reflect significant improvement over sequential architectural design but also consistently outperform leading general-purpose LLMs such as Gemini 2.5 Pro and GPT 4o.

    \item The proposed architecture significantly improves computational efficiency compared to sequential muti-agent architecture, reducing runtime range from 269.2–949.0 seconds to 75.4–194.2 seconds across the benchmark problems. This indicates the effectiveness of the task decomposition strategy in the proposed architecture.
    
    \item The proposed architecture exhibits strong scalability towards larger structural systems. It maintains high accuracy rates of 90\%, 100\%, and 80\% across three large-scale cases involving frames with 7 and 10 bays. It also delivers efficient computation, with inference times of 371.2, 393.0, and 548.6 seconds, respectively. 
    
    \item The proposed architecture demonstrates strong adaptation to diverse linguistic styles. A pilot test involving three students from the School of Architecture at the University of Miami confirms that the architecture consistently produces correct structural analysis results across ten repeated trials, regardless of differences in user input styles.

    \item The proposed architecture utilizes two backbone LLMs to power specialized agents. GPT-OSS 120B is assigned to agents requiring complex reasoning, while Llama-3.3 70B Instruct Turbo is tasked with information mapping and translation. Comparisons against single‑LLM alternatives demonstrate the rationale for this model selection strategy.

    \item The multi-agent LLMs powered by the proposed architecture have been deployed as a publicly accessible web application, enabling community evaluation and real-world testing of multi-step structural modeling and analysis tasks. The website link is \url{https://civilbot.netlify.app}.
    
\end{itemize}

\section*{Data Availability Statement}
Some or all data, models, or code that support the findings of this study are available from the corresponding author upon reasonable request.

\bibliographystyle{ascelike}  
\bibliography{references}

@article{mckenna2011opensees,
  title={OpenSees: a framework for earthquake engineering simulation},
  author={McKenna, Frank},
  journal={Computing in Science \& Engineering},
  volume={13},
  number={4},
  pages={58--66},
  year={2011},
  publisher={IEEE}
}

@manual{etabs2023,
  title        = {{ETABS}: Integrated Building Design Software},
  author       = {{Computers and Structures, Inc.}},
  organization = {Computers and Structures, Inc.},
  address      = {Walnut Creek, CA},
  year         = {2025},
  note         = {\url{https://www.csiamerica.com/products/etabs}},
}

@manual{sap2000,
  title        = {{SAP2000}: Integrated Software for Structural Analysis and Design},
  author       = {{Computers and Structures, Inc.}},
  organization = {Computers and Structures, Inc.},
  address      = {Walnut Creek, CA},
  year         = {2025},
  note         = {url{https://www.csiamerica.com/products/sap2000}},
}

@manual{ansys,
  title        = {{ANSYS}: Engineering Simulation Software},
  author       = {{ANSYS, Inc.}},
  organization = {ANSYS, Inc.},
  address      = {Canonsburg, PA},
  year         = {2025},
  note         = {\url{https://www.ansys.com}},
}

@manual{abaqus,
  title        = {{Abaqus}: Finite Element Analysis},
  author       = {{Dassault Systèmes}},
  organization = {Dassault Systèmes Simulia Corp.},
  address      = {Providence, RI},
  year         = {2025},
  note         = {url{https://www.3ds.com/products-services/simulia/products/abaqus}},
}

@article{an2024make,
  title={Make your llm fully utilize the context},
  author={An, Shengnan and Ma, Zexiong and Lin, Zeqi and Zheng, Nanning and Lou, Jian-Guang and Chen, Weizhu},
  journal={Advances in Neural Information Processing Systems},
  volume={37},
  pages={62160--62188},
  year={2024}
}

@article{zhu2024can,
  title={Can large language models understand context?},
  author={Zhu, Yilun and Moniz, Joel Ruben Antony and Bhargava, Shruti and Lu, Jiarui and Piraviperumal, Dhivya and Zhang, Yuan and Yu, Hong and Tseng, Bo-Hsiang},
  journal={arXiv preprint arXiv:2402.00858},
  year={2024}
}

@article{cheng2025empowering,
  title={Empowering llms with logical reasoning: A comprehensive survey},
  author={Cheng, Fengxiang and Li, Haoxuan and Liu, Fenrong and van Rooij, Robert and Zhang, Kun and Lin, Zhouchen},
  journal={arXiv preprint arXiv:2502.15652},
  year={2025}
}

@article{xie2025logic,
  title={Logic-rl: Unleashing llm reasoning with rule-based reinforcement learning},
  author={Xie, Tian and Gao, Zitian and Ren, Qingnan and Luo, Haoming and Hong, Yuqian and Dai, Bryan and Zhou, Joey and Qiu, Kai and Wu, Zhirong and Luo, Chong},
  journal={arXiv preprint arXiv:2502.14768},
  year={2025}
}

@article{zhou2023instruction,
  title={Instruction-following evaluation for large language models},
  author={Zhou, Jeffrey and Lu, Tianjian and Mishra, Swaroop and Brahma, Siddhartha and Basu, Sujoy and Luan, Yi and Zhou, Denny and Hou, Le},
  journal={arXiv preprint arXiv:2311.07911},
  year={2023}
}

@article{zeng2023evaluating,
  title={Evaluating large language models at evaluating instruction following},
  author={Zeng, Zhiyuan and Yu, Jiatong and Gao, Tianyu and Meng, Yu and Goyal, Tanya and Chen, Danqi},
  journal={arXiv preprint arXiv:2310.07641},
  year={2023}
}

@article{wan2025som,
  title={Som-1k: A thousand-problem benchmark dataset for strength of materials},
  author={Wan, Qixin and Wang, Zilong and Zhou, Jingwen and Wang, Wanting and Geng, Ziheng and Liu, Jiachen and Cao, Ran and Cheng, Minghui and Cheng, Lu},
  journal={arXiv preprint arXiv:2509.21079},
  year={2025}
}

@article{jiang2025large,
  title={Large language model for post-earthquake structural damage assessment of buildings},
  author={Jiang, Yongqing and Wang, Jianze and Shen, Xinyi and Dai, Kaoshan},
  journal={Computer-Aided Civil and Infrastructure Engineering},
  year={2025},
  publisher={Wiley Online Library}
}

@article{xutwo,
  title={A Two-Stage Multi-Modal LLM Fine-Tuning Framework for Analyzing Building Surface Defects},
  author={Xu, Gengyang and Pan, Feng and Yuen, Pong C},
  year={2025}
}

@article{pu2024autorepo,
  title={AutoRepo: A general framework for multimodal LLM-based automated construction reporting},
  author={Pu, Hongxu and Yang, Xincong and Li, Jing and Guo, Runhao},
  journal={Expert Systems with Applications},
  volume={255},
  pages={124601},
  year={2024},
  publisher={Elsevier}
}

@article{joffe2025framework,
  title={The framework and implementation of using large language models to answer questions about building codes and standards},
  author={Joffe, Isaac and Felobes, George and Elgouhari, Youssef and Talebi Kalaleh, Mohammad and Mei, Qipei and Chui, Ying Hei},
  journal={Journal of Computing in Civil Engineering},
  volume={39},
  number={4},
  pages={05025004},
  year={2025},
  publisher={American Society of Civil Engineers}
}

@article{deng2025bimgent,
  title={BIMgent: Towards Autonomous Building Modeling via Computer-use Agents},
  author={Deng, Zihan and Du, Changyu and Nousias, Stavros and Borrmann, Andr{\'e}},
  journal={arXiv preprint arXiv:2506.07217},
  year={2025}
}

@article{du2024text2bim,
  title={Text2BIM: Generating building models using a large language model-based multi-agent framework},
  author={Du, Changyu and Esser, Sebastian and Nousias, Stavros and Borrmann, Andr{\'e}},
  journal={arXiv preprint arXiv:2408.08054},
  year={2024}
}

@article{dong2025ai,
  title={AI BIM coordinator for non-expert interaction in building design using LLM-driven multi-agent systems},
  author={Dong, Yaxian and Zhan, Zijun and Hu, Yuqing and Doe, Daniel Mawunyo and Han, Zhu},
  journal={Automation in Construction},
  volume={180},
  pages={106563},
  year={2025},
  publisher={Elsevier}
}

@article{liu2026large,
  title={A large language model-empowered agent for reliable and robust structural analysis},
  author={Liu, Jiachen and Geng, Ziheng and Cao, Ran and Cheng, Lu and Bocchini, Paolo and Cheng, Minghui},
  journal={Structure and Infrastructure Engineering},
  pages={1--16},
  year={2026},
  publisher={Taylor \& Francis}
}

@article{geng2025lightweight,
  title={A lightweight large language model-based multi-agent system for 2d frame structural analysis},
  author={Geng, Ziheng and Liu, Jiachen and Cao, Ran and Cheng, Lu and Wang, Haifeng and Cheng, Minghui},
  journal={arXiv preprint arXiv:2510.05414},
  year={2025}
}

@article{liang2025integrating,
  title={Integrating large language models for automated structural analysis},
  author={Liang, Haoran and Kalaleh, Mohammad Talebi and Mei, Qipei},
  journal={arXiv preprint arXiv:2504.09754},
  year={2025}
}

@article{sclar2023quantifying,
  title={Quantifying Language Models' Sensitivity to Spurious Features in Prompt Design or: How I learned to start worrying about prompt formatting},
  author={Sclar, Melanie and Choi, Yejin and Tsvetkov, Yulia and Suhr, Alane},
  journal={arXiv preprint arXiv:2310.11324},
  year={2023}
}

@article{chatterjee2024posix,
  title={Posix: A prompt sensitivity index for large language models},
  author={Chatterjee, Anwoy and Renduchintala, HSVNS Kowndinya and Bhatia, Sumit and Chakraborty, Tanmoy},
  journal={arXiv preprint arXiv:2410.02185},
  year={2024}
}

@article{zhu2018openseespy,
  title={OpenSeesPy: Python library for the OpenSees finite element framework},
  author={Zhu, Minjie and McKenna, Frank and Scott, Michael H},
  journal={SoftwareX},
  volume={7},
  pages={6--11},
  year={2018},
  publisher={Elsevier}
}

@misc{kokot_opsvis_2024,
  author       = {Seweryn Kokot},
  title        = {OpsVis Documentation},
  year         = {2024},
  howpublished = {\url{https://sewerynkokot.github.io/opsvis/}},
  note         = {Accessed: 2025-06-07}
}

@misc{openai2025gpt52update,
  title        = {Update to GPT-5 System Card: GPT-5.2},
  author       = {{OpenAI}},
  year         = {2025},
  howpublished = {\url{https://openai.com/index/gpt-5-system-card-update-gpt-5-2/}},
  note         = {Accessed: 2026-01-06}
}

@misc{google2025gemini3procard,
  title        = {Gemini 3 Pro Model Card},
  author       = {{Google DeepMind}},
  year         = {2025},
  howpublished = {\url{https://storage.googleapis.com/deepmind-media/Model-Cards/Gemini-3-Pro-Model-Card.pdf}},
  note         = {Accessed: 2026-01-06}
}

@article{wang2025integrated,
  title={An integrated approach for automatic safety inspection in construction: Domain knowledge with multimodal large language model},
  author={Wang, Yiheng and Luo, Hanbin and Fang, Weili},
  journal={Advanced Engineering Informatics},
  volume={65},
  pages={103246},
  year={2025},
  publisher={Elsevier}
}

@article{shi2025fine,
  title={Fine-tuning a large language model for automated code compliance of building regulations},
  author={Shi, Jack Wei Lun and Solihin, Wawan and Yeoh, Justin KW},
  journal={Advanced Engineering Informatics},
  volume={68},
  pages={103676},
  year={2025},
  publisher={Elsevier}
}
%%% Remove comment to use the external .bib file (using bibtex).
%%% and comment out the ``thebibliography'' section.

%%% Comment out this section when you \bibliography{references} is enabled.
% \begin{thebibliography}{1}

% \bibitem{kour2014real}
% George Kour and Raid Saabne.
% \newblock Real-time segmentation of on-line handwritten arabic script.
% \newblock In {\em Frontiers in Handwriting Recognition (ICFHR), 2014 14th
%   International Conference on}, pages 417--422. IEEE, 2014.

% \end{thebibliography}

\clearpage
\appendix
\section*{Appendix: Hallucination patterns in multi-agent architectures powered by single LLMs}
\label{appendix:translation and validation}

\renewcommand{\thefigure}{A.\arabic{figure}}  % Change figure numbering to A.1, A.2, etc.
\setcounter{figure}{0}

\vspace{2em}

\begin{figure*}[htbp]
\centering
% \captionsetup{justification=centering}
\includegraphics[width=0.85\textwidth]{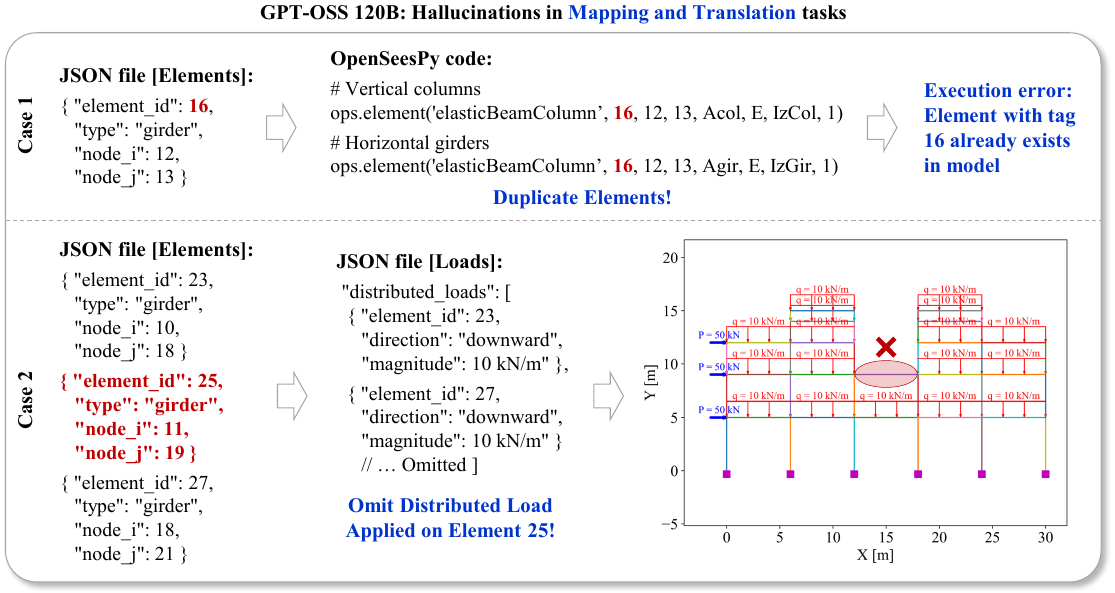}
\caption{GPT-powered system: hallucinations concentrated in mapping and translation tasks..}
\label{fig:A1}
\end{figure*}

\vspace{2em}

\begin{figure*}[htbp]
\centering
% \captionsetup{justification=centering}
\includegraphics[width=0.85\textwidth]{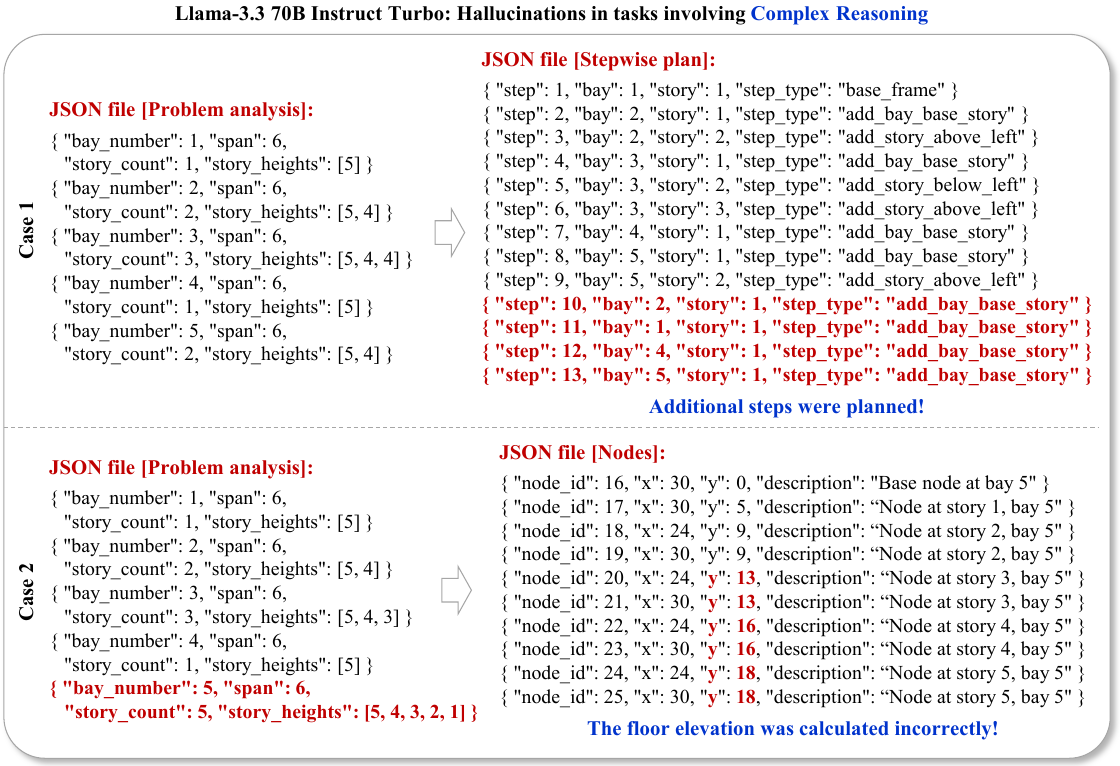}
\caption{Llama-powered system: hallucinations arising during complex reasoning tasks.}
\label{fig:A2}
\end{figure*}

\clearpage
\vspace*{-0.5cm} 
\begin{figure*}[t]
\centering
% \captionsetup{justification=centering}
\includegraphics[width=0.85\textwidth]{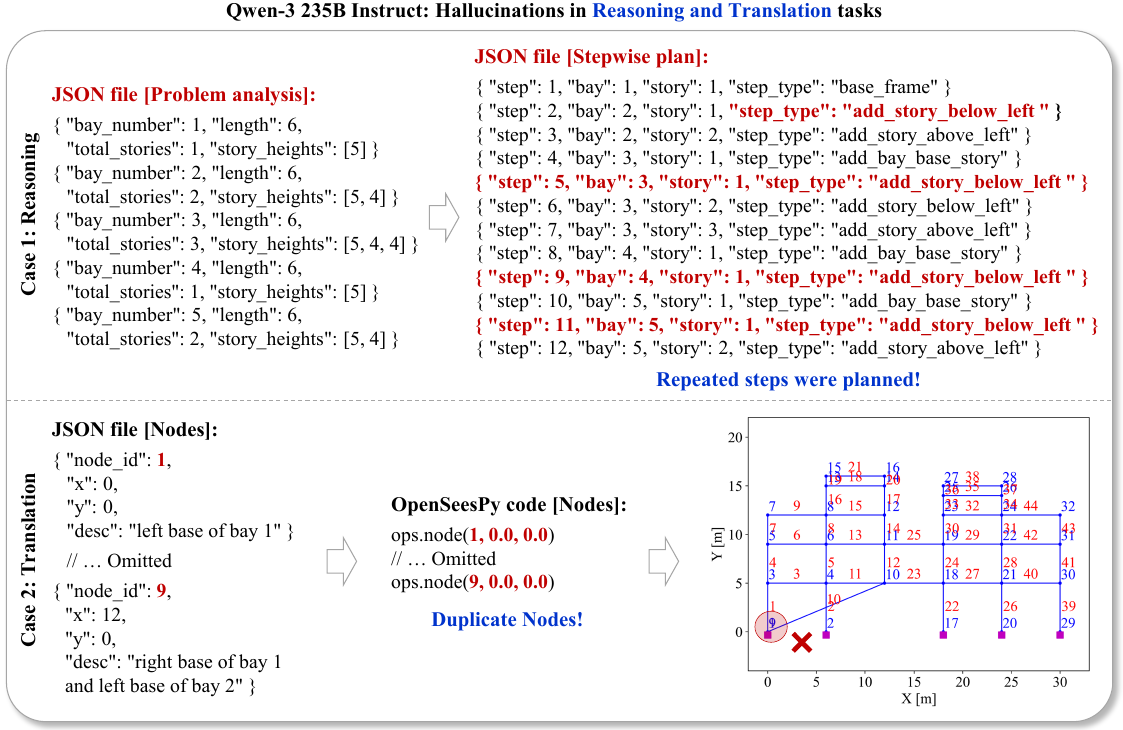}
\caption{Qwen-powered system: hallucinations involving both reasoning and translation tasks.}
\label{fig:A3}
\end{figure*}

\end{document}